\documentclass[10pt,conference,letterpaper]{IEEEtran}

\ifCLASSOPTIONcompsoc
  \usepackage[nocompress]{cite}
\else
  \usepackage{cite}
\fi

\usepackage[ruled,vlined]{algorithm2e}
\usepackage{array}
\usepackage{amsfonts}
\usepackage{amsmath}
\usepackage{amssymb}
\usepackage{booktabs}
\usepackage{caption}
\usepackage{color}
\usepackage{comment}
\usepackage[inline,shortlabels]{enumitem}
\usepackage[pdftex]{graphicx}
\usepackage{epstopdf}
\usepackage{newtxmath}
\usepackage{subcaption}
\usepackage{url}

\usepackage{breakurl}
\usepackage[bookmarks=false]{hyperref}
\usepackage{cuted}
\usepackage{flushend}

\graphicspath{ {figures/} }

\newcommand{\eg}{\emph{e.g.},\xspace}

\newcommand{\myparagraph}[1]{\vspace{0.5em}\noindent\textbf{#1}}
\newcommand{\ie}{\emph{i.e.,}\xspace}

\newcommand{\etal}{\emph{et~al.}\xspace}
\newcolumntype{x}[1]{>{\raggedright\arraybackslash}p{#1}}

\SetCommentSty{mycommfont}

\DeclareMathOperator*{\argmax}{argmax}
\DeclareMathOperator*{\argmin}{argmin}

\makeatletter 
\g@addto@macro{\@algocf@init}{\SetKwInOut{Output}{Output}} 
\makeatother

\begin{document}

\title{Adversarial Reinforcement Learning under Partial Observability in Autonomous Computer Network Defence}

\author{\IEEEauthorblockN{Yi~Han\IEEEauthorrefmark{1}, David~Hubczenko\IEEEauthorrefmark{2}, Paul~Montague\IEEEauthorrefmark{2}, Olivier~De Vel\IEEEauthorrefmark{2}, Tamas~Abraham\IEEEauthorrefmark{2},\\ Benjamin~I.~P.~Rubinstein\IEEEauthorrefmark{1}, Christopher~Leckie\IEEEauthorrefmark{1}, Tansu~Alpcan\IEEEauthorrefmark{3}, Sarah~Erfani\IEEEauthorrefmark{1}}
\IEEEauthorblockA{\IEEEauthorrefmark{1}School of Computing and Information Systems\\ The University of Melbourne, Parkville, Australia, 3010\\
Email: \{yi.han, benjamin.rubinstein, caleckie, sarah.erfani\}@unimelb.edu.au}
\IEEEauthorblockA{\IEEEauthorrefmark{2}Defence Science and Technology Group, Australian Department of Defence\\ Edinburgh, Australia, 5111\\
Email: \{david.hubczenko, paul.montague, olivier.devel, tamas.abraham\}@dst.defence.gov.au}
\IEEEauthorblockA{\IEEEauthorrefmark{3}Department of Electrical and Electronic Engineering\\ The University of Melbourne, Parkville, Australia, 3010\\
Email: tansu.alpcan@unimelb.edu.au}}

\maketitle

\begin{abstract}
Recent studies have demonstrated that reinforcement learning (RL) agents are susceptible to adversarial manipulation, similar to vulnerabilities previously demonstrated in the supervised learning setting. While most existing work studies the problem in the context of computer vision or console games, this paper focuses on reinforcement learning in autonomous cyber defence under partial observability. We demonstrate that under the black-box setting, where the attacker has no direct access to the target RL model, causative attacks---attacks that target the training process---can poison RL agents even if the attacker only has partial observability of the environment. In addition, we propose an inversion defence method that aims to apply the opposite perturbation to that which an attacker might use to generate their adversarial samples. Our experimental results illustrate that the countermeasure can effectively reduce the impact of the causative attack, while not significantly affecting the training process in non-attack scenarios.
\end{abstract}


\section{Introduction}\label{sec:intro}
The adversarial machine learning~\cite{barreno_security_2010,huang2011adversarial,biggio_poisoning_2012,szegedy_intriguing_2013} literature has demonstrated that machine learning models are vulnerable to both exploratory (test-time) and causative (training-time) attacks. These attacks are typically crafted by applying calculated perturbations to the test or training instances, in order to either cause misclassification or poison the training process. More recent studies~\cite{huang_adversarial_2017,behzadan_vulnerability_2017,han_reinforcement_2018,pattanaik_robust_2017} have shown that similar attacks can also be effective against reinforcement learning algorithms.

Unlike the majority of the literature that mainly focuses on the vision domain or console games, in previous work we \cite{han_reinforcement_2018} analyse how reinforcement learning agents react to different forms of poisoning attacks in the context of autonomous defence in software-defined networking (SDN)~\cite{noauthor_sdn_2014}. Specifically, we first demonstrate that without any poisoning attacks, an RL agent can be successfully trained to identify the optimal strategy for preventing the attacker from propagating through the network. Then we investigate the effect of two different types of poisoning attacks on the RL training process, and show that the RL agent can be misled into making non-optimal decisions, causing a significantly larger part of the network to be compromised by the attacker. Section \ref{sec:problem} provides a more detailed description. 

However, there are two limitations with the previous work~\cite{han_reinforcement_2018}: (1) full observability of the (network) states is assumed in the analysis, which is often not the case in real-world situations, especially for the attacker; (2) while an important topic, treatment of RL defence mechanisms is preliminary, and the proposed method does not work effectively in the new setup as introduced below. In this work, we address these limitations and make the following \textbf{\emph{contributions}}: 

First, we impose \textbf{\emph{partial observability}} for the attacker. Since it is unlikely that the attacker can map out the entire network topology, we consider the scenario where the defender has full observability of the network, but the attacker only knows part of the topology. Specifically, Fig.~\ref{figure_network} depicts the running example network studied in this paper. The network is comprised of 100 nodes and 172 links, and the attacker has an initial foothold of a handful of compromised nodes. They aim to propagate through the network to take control of a specific node corresponding to the critical server, which in response can be migrated by the defender to some pre-determined alternate nodes.

As shown in the figure, two setups are considered, where the attacker can observe around one-third and half of all the nodes, respectively. Under each setup, the defender trains a reinforcement learning agent to (1) protect the critical server from being compromised, and (2) maintain the network functionality as much as possible, \ie maximise the number of nodes that can reach the critical server. On the other hand, the attacker only has partial observability, which restricts their action set: they cannot compromise an adjacent node unless the link to the node is known.

\begin{figure*}[ht!]
\centering
\includegraphics[width=.9\textwidth]{networks.pdf}
\caption{Network setups: (a) the attacker can observe 34 nodes and 46 links; (b) the attacker can observe 51 nodes and 80 links.}
\label{figure_network}
\end{figure*}

Second, we propose a new \textbf{\emph{inversion defence method}} to counteract the causative attack on reinforcement learning algorithms. Our experimental results suggest that the approach introduced in~\cite{han_reinforcement_2018} does not work well in our setup (Fig.~\ref{figure_network}). Instead, we design a method that requires no prior knowledge about the attacker, and attempts to undo attacker poisoning of the RL training process. We demonstrate the effectiveness of the new defensive algorithm, and show that it has limited impact in non-attack scenarios.

The remainder of this paper is organised as follows: 
Section~\ref{sec:problem} summarises the problem of applying reinforcement learning for autonomous defence in computer networks; 
Section~\ref{sec:attack} introduces the causative attack via state perturbation and Section~\ref{sec:defence} the defence mechanism; 
Section~\ref{sec:experiment} presents the experimental verification; 
Section~\ref{sec:related} reviews previous work in adversarial machine learning; 
and finally Section~\ref{sec:conc} concludes the paper and offers directions for future work.

\section{Problem: Reinforcement Learning for Autonomous Network Defence}
\label{sec:problem}
We now overview the problem of autonomous defence in computer networks using reinforcement learning. 

\subsection{Background on Reinforcement Learning}
Reinforcement learning~\cite{sutton_introduction_1998} deals with a sequential decision making problem where an agent interacts with the environment to maximise its rewards. At each time step \(t\), the agent (1) receives an observation \(s_{t}\) of the environment; (2) takes an action \(a_{t}\) based on its policy \(\pi\), which is a mapping from states to actions; 
and (3) obtains a reward \(r_{t}\) based on state $s_t$, action $a_t$, and the environment's transition to a new state \(s_{t+1}\). 
The goal of the agent is to maximise its cumulative rewards, \ie \(R_{t} = \sum_{\tau=t}^{\infty} \gamma^{\tau-t}r_{\tau}\), where \(\gamma \in (0, 1]\) is a discount factor which affects the present importance of long-term rewards.

We focus our experiments on two widely used RL algorithms---Double Deep Q-Networks (DDQN)~\cite{van_hasselt_deep_2015} and Asynchronous Advantage Actor-Critic (A3C)~\cite{mnih_asynchronous_2016}---and the transferability of attacks between them.

\subsection{Autonomous Network Defence with Reinforcement Learning}
Maintaining the security of cyber environments without affecting the normal exchange of information is a challenging task. Although the problem of cyber defence has been studied for decades, most deployed solutions are still rule-based that require a significant human involvement and are prone to generating false alarms---rules are formulated based on previously seen threats, and may not be applicable to new vulnerabilities. In addition, for those solutions that do employ machine learning, traditional one-class or two-class prediction algorithms are often used, which require prior knowledge on existing attacks to make accurate decisions. However, acquiring the prior knowledge on all existing attacks is almost impossible, as more sophisticated attacks are generated everyday.

In this work, we investigate the feasibility of applying RL for autonomous cyber defence, as RL has the ability to adapt and generalise, and has been successfully used for autonomous control in a wide range of applications.

Specifically, we consider a computer network of \(|N|\) nodes, \(N=\{n_{1}, n_{2}, ..., n_{|N|}\}\), and \(|L|\) links, \(L=\{l_{1}, l_{2}, ..., l_{|L|}\}\) (\eg Fig.~\ref{figure_network}), where \(N_{D} \subset N\) is the set of critical servers to be protected (one or more blue nodes), \(N_{M} \subset N\) is the set of possible migration destinations for node \(n \in N_{D}\) (one or more green nodes), and \(N_{A} \subset N\) is the set of nodes that have been compromised (red nodes). In addition, while the defender knows all the nodes and links, the attacker is only able to map out a subset of them---\(N_{O}\) and \( L_{O}\), where \(N_{O} \subseteq N, L_{O} \subseteq L\).

The attack scenario we consider is a cyber attack against the network infrastructure. Here, the attack spreads through the network, and aims to take control of the critical servers (note that we assume that the attacker has to compromise all nodes on the path). However, they can compromise a node \(n\) only if there is a link \(l \in L_{O}\) between \(n\) and a compromised node \(n^{\prime} \in N_{A}\). That is \(N_{A}\) keeps expanding as the attack proceeds.

In order to protect the critical servers from being compromised, the defender trains an RL agent that:
\begin{enumerate}[align=left, leftmargin=0pt, labelindent=0pt, listparindent=0pt, labelwidth=0pt, itemindent=!]
    \item Monitors the system state. The system state is represented using a binary feature representation. The state representation has a number of bits equal to the sum of the number of nodes and number of links in the network. A bit corresponding to a node is 0/1 to represent whether that node is un/compromised. A bit corresponding to a link is 0/1 to represent whether that link is down/up. Note that \textbf{detection is not our focus}---we are not studying how to detect the attacker, nor how the attacker can escape detection. Therefore, we have modeled the defender as having in place a detection system. Our experiments suggest that as long as the system achieves a reasonable detection rate, e.g., \(\geq 75\%\), it does not have an obvious impact on the final results. In our experiment, the detection rate is set to 90\%.
    \item Chooses an appropriate action to take when in a given system state. The actions that are available comprise: (i) isolating and patching one node; (ii) reconnecting one node and its links; (iii) migrating the critical server to a certain destination; and (iv) taking no action. Note that for actions (i) or (ii), only one node can be isolated or reconnected during each action cycle.
\end{enumerate}

The reward function that the RL agent is trained on is given in (\ref{eq_reward}), where \(U_{t}\) is the number of nodes unreachable from the critical server after the current \(t^{th}\) step, \(C_{t}\) is the number of newly compromised nodes at the \(t^{th}\) step, \(r_{c}\) is the reward for an additional node to be compromised, \(r_{m}\) is the migration cost, \(\boldsymbol{1_{a=m}} = 1\) iff the action \(a\) is to migrate the critical server, and \(\alpha,\ \beta \geq 1.0\).
\begin{equation}\label{eq_reward}
    r_{t} = \left\{
                \begin{array}{l}
                     -1,\ \ n \in N_{D} \text{ is compromised or the action is invalid}\\
                     \left(1 - \alpha \cdot \frac{U_{t}}{|N|} \cdot \beta^{t}\right) - C_{t} \cdot r_{c} \cdot \beta^{t} - \boldsymbol{1_{a=m}} \cdot r_{m}, \  \text{otherwise}
                \end{array}
        \right.
\end{equation}

As we can see, the reward is based on (i) whether any critical server has been compromised; (ii) the validity of an action, \eg if a node has already been isolated, it cannot be isolated again; (iii) the number of nodes reachable from the critical servers; (iv) the number of newly compromised nodes; and (v) the migration cost. Note that the term \(\beta^{t}\) encourages RL agents to find the optimal solution with minimal steps.

\begin{figure*}[t!]
\centering
\includegraphics[width=.9\textwidth]{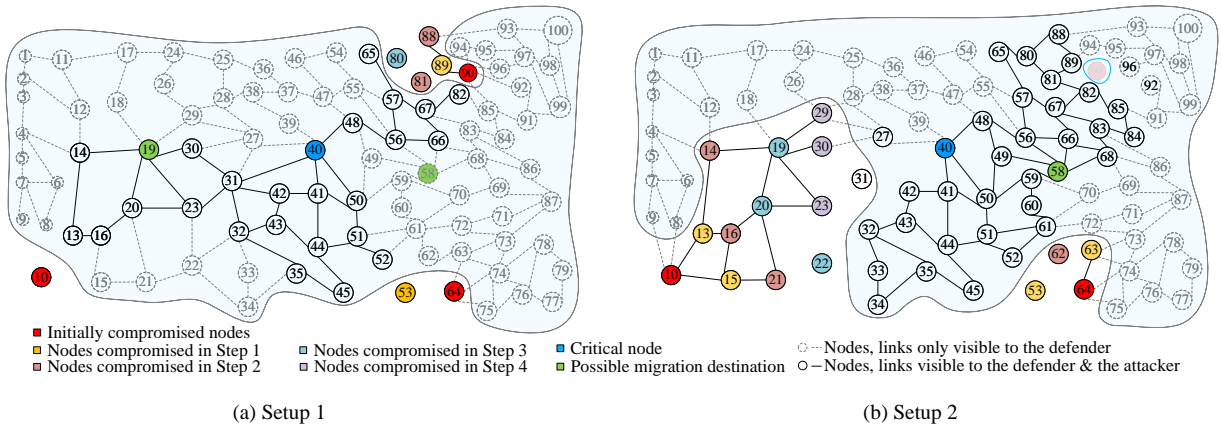}
\caption{Optimal results in response to a cyber attack against the network (in the absence of attacking the RL algorithm).}
\label{figure_solution}
\end{figure*}

For each of the setups in Fig.~\ref{figure_network}, we train multiple DDQN (with Prioritised Experience Replay~\cite{schaul_prioritized_2015}) and A3C agents with different structures, i.e., different numbers of hidden layers and different numbers of neurons per layer. These agents help us identify the optimal policy without tampering (see Fig.~\ref{figure_solution}): (1) under the first setup, isolating nodes in the order of 10, 53, 81, 80, which results in a total of 92 out of 100 nodes being preserved. Note that there are several other equally optimal solutions for this case; (2) under the second setup, isolating nodes in the order of 90, 53, 62, 22, 31, which results in a total of 82 out of 100 nodes being preserved.

However, the above cyber attack scenario and resulting trained RL agents leave important questions unanswered: \textbf{\emph{if the attacker has the ability to poison the training process, can the agents still identify the optimal actions? What can the defender do to mitigate attack impact?}} We seek to address these questions.

\section{Partially-Observable Poisoning Attacks on RL by State Manipulation}\label{sec:attack}
In order for RL techniques to be successfully applied in autonomous cyber defence, it is crucial to analyse the susceptibility of RL agents to potential causative attacks. However, most existing adversarial attacks against RL agents are based on gradient descent optimisation~\cite{huang_adversarial_2017,behzadan_vulnerability_2017,lin_tactics_2017,pattanaik_robust_2017}, and in our case the attacker aims to manipulate the binary state of a node (note again that the purpose of the attack is not to escape detection/cause misclassification). Therefore, gradient descent-based attacks are not applicable. Instead, we have investigated the following attack mechanisms: 

\begin{enumerate}[align=left, leftmargin=0pt, labelindent=0pt, listparindent=0pt, labelwidth=0pt, itemindent=!]
    \item Tampering with a small number (\eg 5\%) of rewards to maximise the defender's loss. Specifically, the gradient of the loss with respect to the rewards, is used to select which rewards to tamper with;
    \item Random perturbation of the observed states;
    \item Manipulating the states to minimise the defender's rewards;
    \item Manipulating the states to minimise the probability of taking the optimal action.
\end{enumerate}

In our preliminary unreported experiments we found that the last attack mechanism was the most effective and hence we subsequently use it as the attacker's strategy.

\subsection{Threat Model}
We focus on the scenario where the attacker tampers with the states observed by the RL agents, so that the trained model learns sub-optimal actions. Specifically, suppose that the agent observes an experience \((s, a, s^{\prime}, r)\) without any attacks, where \(s\) is the current system state, \(a\) is the action taken by the agent, \(s^{\prime}\) is the new state, and \(r\) is the reward. When the system reaches the new state \(s^{\prime}\), the agent would continue to take the next optimal action \(a^{\prime}\). The attacker can counteract this by introducing false positive (FP) and false negative (FN) readings in \(s^{\prime}\), meaning that uncompromised (compromised) nodes will be reported as compromised (uncompromised) to the defender. Consequently, the agent observes \((s, a, s^{\prime}+\delta, r^{\prime})\) (where \(\delta\) represents the FP and FN readings) instead of \((s, a, s^{\prime}, r)\), and hence may not take action \(a^{\prime}\) next.

The key issue here is how the attacker chooses the nodes to manipulate. We consider the following strategy:
\begin{enumerate}[align=left, leftmargin=0pt, labelindent=0pt, listparindent=0pt, labelwidth=0pt, itemindent=!]
    \item Against the DDQN agent: loop through all observable nodes to find \(\delta\) that minimises the \(Q\)-value of the optimal action \(a^{\prime}\) for state \(s^{\prime}+\delta\), \ie \(\argmin_{\delta} Q(s^{\prime}+\delta, a^{\prime})\);
    \item Against the A3C agent: loop through all observable nodes to find \(\delta\) that minimises the probability of taking the optimal action \(a^{\prime}\) for state \(s^{\prime}+\delta\), \ie \(\argmin_{\delta} \pi(a^{\prime}|s^{\prime}+\delta)\).
\end{enumerate}

We next abstract the threat model for adversarial learning in autonomous cyber defence as follows:

\myparagraph{Black-box approach.} The attacker does not have access to the defender's training model as per our partial observability assumption. This constitutes a form of black-box attack, which means the attacker needs to train their own surrogate model first, based on the partial topology visible to them.

\myparagraph{Limited choice of potential false positive and false negative nodes.} It is unlikely that the attacker can falsify the state of all observable nodes. Therefore, we limit the nodes whose states can be perturbed by the attacker. Section~\ref{sec:experiment} further explains how these nodes are selected.

\myparagraph{Limits on the number of false readings per time step.} In our experiments, the number of FP and FN nodes that can be introduced per time step are no more than two per case. 

Our view is that this model of attacker information/control is a key point of interest in exploring domains beyond computer vision. Algorithm~\ref{algo:causative_att} details this attack against DDQN. The algorithm for attacks against A3C is similar and so is omitted.

\begin{algorithm}[t!]
\LinesNumbered
\SetKwInOut{Input}{\small Input}
\SetKwInOut{Output}{\small Output}
\Input{The original experience, \((s, a, s^{\prime}, r)\); \\
The list of observable nodes, \(N_{O}\); \\
The list of nodes that can be perturbed as \\false positive (false negative) by the attacker, \\ \(L_{\mathit{FP}}\) (\(L_{\mathit{FN}}\)); \\
The main DQN, \(Q\); \\
Limit on the number of FPs and FNs per \\time, LIMIT}
\Output{The tampered experience \((s, a, s^{\prime}+\delta, r^{\prime})\)}
\BlankLine

\(\mathit{FN} = \mathit{FP} = \{\}\)\;
\(minQ_\mathit{FN} = minQ_\mathit{FP} = \{\}\)\;
\(a^{\prime} = \argmax_{a^{*}} Q(s^{\prime}, a^{*})\)\;
\For{node \(n\) in \(N_{O}\)}{
	\If{\(n\) is compromised and \(n\) in \(L_\mathit{FN}\)}{
		mark \(n\) as uncompromised\;
		\If{\(Q(s^{\prime}+\delta, a^{\prime}) <\) any value in \(minQ_\mathit{FN}\)}{
			\tcp{\(\delta\) represents the FP and/or FN readings}
			insert \(n\) and \(Q(s^{\prime}+\delta, a^{\prime})\) into appropriate positions in \(FN\) and \(minQ_\mathit{FN}\)\;
			\If{\(|FN| > LIMIT\)}{
			    remove extra nodes from \(FN\) and \(minQ_\mathit{FN}\)\;
			}
		}
		restore \(n\) as compromised\;
	}
	\ElseIf{\(n\) is uncompromised and \(n\) in \(L_\mathit{FP}\)}{
		mark \(n\) as compromised\;
		\If{\(Q(s^{\prime}+\delta, a^{\prime}) <\) any value in \(minQ_\mathit{FP}\)}{
			insert \(n\) and \(Q(s^{\prime}+\delta, a^{\prime})\) into appropriate positions in \(FP\) and \(minQ_\mathit{FP}\)\;
			\If{\(|FP| > LIMIT\)}{
			    remove extra nodes from \(FP\) and \(minQ_\mathit{FP}\)\;
			}
		}
		restore \(n\) as uncompromised\;
	}
}
Change nodes in \(FN\) to uncompromised\;

Change nodes in \(FP\) to compromised\;

\Return{\((s, a, s^{\prime}+\delta, r^{\prime})\)}

\caption{Causative attack against DDQN via state perturbation\label{algo:causative_att}}
\end{algorithm}

\section{The Inversion Defence Mechanism}\label{sec:defence}
For the defender we aim to design a defence mechanism that (1) effectively mitigates the impact of the above causative attack, (2) requires minimum knowledge of the attacker, and (3) does not affect the training when there is no attack. Specifically, we propose a countermeasure that generates training instances by applying a perturbation counter to simulated adversarial samples.

Since the attacker adds false readings \(\delta\) into the observed states, can \(\delta\) be reversed? If the defender knows the nodes that are visible to the attacker, limits on the FP \& FN nodes, and the number of FPs and FNs added per time step, then they may find these false readings, by solving the inverse problem of how the attacker generates the adversarial samples: while the attacker receives \((s, a, s^{\prime}, r)\), and loops through all observable nodes to find \(\delta\) that either minimises the \(Q\)-value \(Q(s^{\prime}+\delta, a^{\prime})\) or the probability \(\pi(a^{\prime}|s^{\prime}+\delta)\) of action \(a^{\prime}\) for state \(s^{\prime}+\delta\), the defender receives \((s, a, s^{\prime}+\delta, r^{\prime})\), and searches within the same nodes to find \(\delta^{\prime}\) that maximises \(Q(s^{\prime}+\delta+\delta^{\prime}, a^{\prime})\) for DDQN, and \(\pi(a^{\prime}|s^{\prime}+\delta+\delta^{\prime})\) for A3C. In other words, \(\delta^{\prime} = -\delta\).

However, the defender does not know (1) the attacker's partial knowledge of the network topology, (2) the limits on the choice of FP \& FN nodes, and (3) the number of false readings per time interval/step. As shown in Algorithm~\ref{algo:causative_def}, we propose the following to address these obstacles:

\begin{enumerate}[align=left, leftmargin=0pt, labelindent=0pt, listparindent=0pt, labelwidth=0pt, itemindent=!]
    \item Instead of looping through the nodes observable to the attacker, the defender necessarily goes through all network nodes to find \(\delta^{\prime}\)---this solves the first two issues, but increases the training time. We further discuss the overhead in Section~\ref{subsec:discussion};
    \item Test the scenarios where compared with the actual number of false readings introduced by the attacker at each time step, the defender assumes less, the same and more added---as demonstrated in our experiments (Section~\ref{sec:experiment_defence}), even if the defender does not know the exact number of false readings, the inversion defence method is still effective.
\end{enumerate}

\(\delta^{\prime}\) obtained in such a way may not exactly match \(\delta\), and the defender can choose to either keep both \((s, a, s^{\prime}+\delta, r^{\prime})\) and \((s, a, s^{\prime}+\delta+\delta^{\prime}, r^{\prime})\), or only the latter. This method does not make any assumptions about the attacker, except that they falsify the states of certain nodes. However, as demonstrated by the results in Section~\ref{sec:experiment}, the method is effective against the causative attack, and it does not prevent the agent from learning the optimal actions in the non-attack scenario.

\begin{algorithm}[t!]
\LinesNumbered
\SetKwInOut{Input}{\small Input}
\SetKwInOut{Output}{\small Output}
\Input{The potentially tampered experience, \((s, a, s^{\prime}+\delta, r^{\prime})\); \\
The main DQN, \(Q\); \\
The list if all nodes, \(N\); \\
The estimate of the attacker's limit on the \\number of FPs and FNs per time, \(LIMIT^{\prime}\)}
\Output{The corrected experience \((s, a, s^{\prime}+\delta+\delta^{\prime}, r^{\prime})\)}
\BlankLine

\(FN = FP = \{\}\); \tcp{\(FN(FP)\) is a list of potentially false negative (false positive) nodes tampered by the adversaries that need to be corrected}
\(maxQ_\mathit{FN} = maxQ_\mathit{FP} = \{\}\)\;
\(a^{\prime} = \argmax_{a^{*}} Q(s^{\prime}+\delta, a^{*})\)\;
\For{node \(n\) in \(N\)}{
	\If{\(n\) is compromised}{
		mark \(n\) as uncompromised\;
		\If{\(Q(s^{\prime}+\delta+\delta^{\prime}, a^{\prime}) >\) any value in \(maxQ_\mathit{FP}\)}{
		    \tcp{\(\delta^{\prime}\) represents the correction introduced by the defender}
		    \tcp{\(n\) is potentially a false positive node}
			insert \(n\) and \(Q(s^{\prime}+\delta+\delta^{\prime}, a^{\prime})\) into appropriate positions in \(FP\) and \(maxQ_\mathit{FP}\)\;
			\If{\(|FP| > LIMIT^{\prime}\)}{
			    remove extra nodes from \(FP\) and \(maxQ_\mathit{FP}\)\;
			}
		}
		restore \(n\) as compromised\;
	}
	\ElseIf{\(n\) is uncompromised}{
		mark \(n\) as compromised\;
		\If{\(Q(s^{\prime}+\delta+\delta^{\prime}, a^{\prime}) >\) any value in \(maxQ_\mathit{FN}\)}{
		    \tcp{\(n\) is potentially a false negativnode}
			insert \(n\) and \(Q(s^{\prime}+\delta+\delta^{\prime}, a^{\prime})\) into appropriate positions in \(FN\) and \(maxQ_\mathit{FN}\)\;
			\If{\(|FN| > LIMIT^{\prime}\)}{
			    remove extra nodes from \(FN\) and \(maxQ_\mathit{FN}\)\;
			}
		}
		restore \(n\) as uncompromised\;
	}
}
Change nodes in \(FN\) to compromised\;
Change nodes in \(FP\) to uncompromised\;
\Return{\((s, a, s^{\prime}+\delta+\delta^{\prime}, r^{\prime})\)}

\caption{The inversion defence mechanism\label{algo:causative_def}}

\end{algorithm}

\section{Experimental Results}\label{sec:experiment}
We next introduce our experimental setup, present how DDQN and A3C agents are affected by causative attacks, and demonstrate effectiveness of the proposed defence. 

\subsection{SDN Experimental Environment}
In order to better cope with today's dynamic and high-bandwidth traffic, software-defined networking (SDN)~\cite{noauthor_sdn_2014} is designed as a next-generation tool chain for computer network management. SDN adopts a three layer architecture: (1) in the top application layer, applications that deliver services communicate their network requirements to the controller; (2) in the middle layer, the SDN controller translates the received requirements into low-level controls, and passes them to the bottom infrastructure layer; (3) the infrastructure layer includes switches that control forwarding and data processing. Under such an architecture, the controller has a centralised view of the whole network, and is directly programmable since network control is decoupled from forwarding functions. It is thus convenient to monitor and reconfigure network resources.

There have been a number of proprietary and open-source SDN controller software platforms. In this paper, we choose OpenDaylight~\cite{medved_opendaylight:_2014}, the most popular open-source SDN controller available. 
Specifically, we use Mininet~\cite{noauthor_mininet:_2017}, a popular network emulator, to create the network with 100 nodes and 172 links as shown in Fig.~\ref{figure_network}. Once the network is created, OpenDaylight is added as the controller. It provides APIs for the RL agent to retrieve network information and execute different types of operations as defined in Section~\ref{sec:problem}.

We want to emphasise that \emph{\textbf{SDN is only one platform we choose for demonstration purposes}---although it is used in production. The studied causative attacks and the proposed defence method are not coupled to any particular platform}. 

\subsection{Causative Attacks via State Perturbation}\label{sec:causative_attack}
As described in Section~\ref{sec:attack}, we are considering a black-box setting, which means that the attacker does not have direct access to the target RL model, and needs to train its own model. For each of the setups in Fig.~\ref{figure_network}, we achieve this by training a DDQN agent using the partial topology visible to the attacker. The model is then used as the surrogate to attack both of the defender's models (\ie both DDQN and A3C agents).

In addition, there is a limit on the nodes that the attacker can perturb. This is an appropriate threat model---even if the attacker can map out part of the network topology, it is very unlikely that they can manipulate the states of all those nodes. We run the attack by adding one FP and one FN per time interval/step but without any limits on the choices of FPs and FNs. In this way, we are able to find the nodes that are most frequently selected as FPs and FNs. \(L_\mathit{FP}\) and \(L_\mathit{FN}\) in Algorithm~\ref{algo:causative_att} are then initialised with these nodes. Note that the nodes in \(L_\mathit{FP}\) and \(L_\mathit{FN}\) are different under the two setups, and within each setup they are also different for the DDQN and A3C agents. The attacker is only allowed to manipulate the states of these nodes.

Furthermore, the attacker also needs to limit the number of false positive and false negative readings added per time interval. Considering the practicality of the attack, two settings are used in our experiments: (i) one FP \& one FN, and (ii) two FPs \& two FNs.

Fig.~\ref{figure_attack} shows the effectiveness of the attack under different settings, where the top four, six, eight FP nodes and top two FN nodes are selected, \ie \(|L_\mathit{FP}|=4, 6 \text{ or } 8, \text{ while } |L_\mathit{FN}|=2\). \(|L_\mathit{FN}|\) is set to \(2\) because additional experiments with multiple combinations suggests that further increasing \(|L_\mathit{FN}|\) does not have an obvious impact. The results demonstrate that: 
\begin{enumerate}[align=left, leftmargin=0pt, labelindent=0pt, listparindent=0pt, labelwidth=0pt, itemindent=!]
    \item The causative attack designed in Algorithm~\ref{algo:causative_att} is effective against both DDQN and A3C agents when there is no form of defence---under both setups a significant percentage of attacks either cause the critical server to be compromised (the red bars), or cause fewer nodes to be preserved (the blue bars). Note that this also demonstrates the existence of transferability between RL algorithms \cite{papernot_transferability_2016}---attackers do not need to have knowledge of the defender's model and hence attempting to keep the model secret is not an effective countermeasure against adversarial reinforcement learning attacks.
    \item Under the second setup where the attacker observes more nodes, the attacks are more effective in general---the average number of preserved nodes is much lower in most cases. This is because the effectiveness of the attack depends on how close the surrogate and target models are, and with a larger observable topology, the attacker is more likely to train a surrogate that resembles the target RL agent.
    \item Given the same number of false readings per time step, the stricter the limits on the choices of FPs and FNs, \ie the smaller \(|L_\mathit{FP}|\) and \(|L_\mathit{FN}|\) are, the less powerful the attacks are---not only do the limits restrict which nodes can be manipulated, they also decrease the number of steps that are poisoned in each training episode.
    \item Interestingly, if we compare the second and fourth bars in all four figures, when \(|L_\mathit{FP}|=6\), adding one FP \& one FN per time step is more effective than adding two FPs \& two FNs per time step. This is because more training steps are likely to be poisoned in the former case given that \(|L_\mathit{FP}|\) is the same.
\end{enumerate}

In the next section, we test our proposed countermeasure against the most powerful form of attack as illustrated in Fig.~\ref{figure_attack}, where \(|L_\mathit{FP}|=8\), \(|L_\mathit{FN}|=2\), and two FPs \& two FNs are added per time step under the second setup.

\subsubsection{Discussion on the attack efficiency}
The limited choice of potential false positive and false negative nodes, \ie \(L_\mathit{FP}\) and \(L_\mathit{FN}\), not only makes the attack more practical but also increases the efficiency of the attack, as the attacker only needs to loop through these two lists of nodes to find the FPs and FNs instead of checking all the visible nodes. Our experimental results suggest that the attack does not cause an obvious delay to the normal training process.

\subsubsection{Discussion on the Impact of Partial Observability}
As we mentioned earlier, a subset of nodes are more frequently selected as FPs and FNs. 
Therefore, the attack will become more effective if the attacker can take control over more of these most damaging nodes. For future work, we intend to further study the relation between partial observability and attack effectiveness. Specifically, we will identify a minimum set of nodes that the attacker needs to control for a given level of efficiency.

\begin{figure}[t!]
\centering
\begin{subfigure}{.75\columnwidth}
  \centering
  \includegraphics[width=\columnwidth]{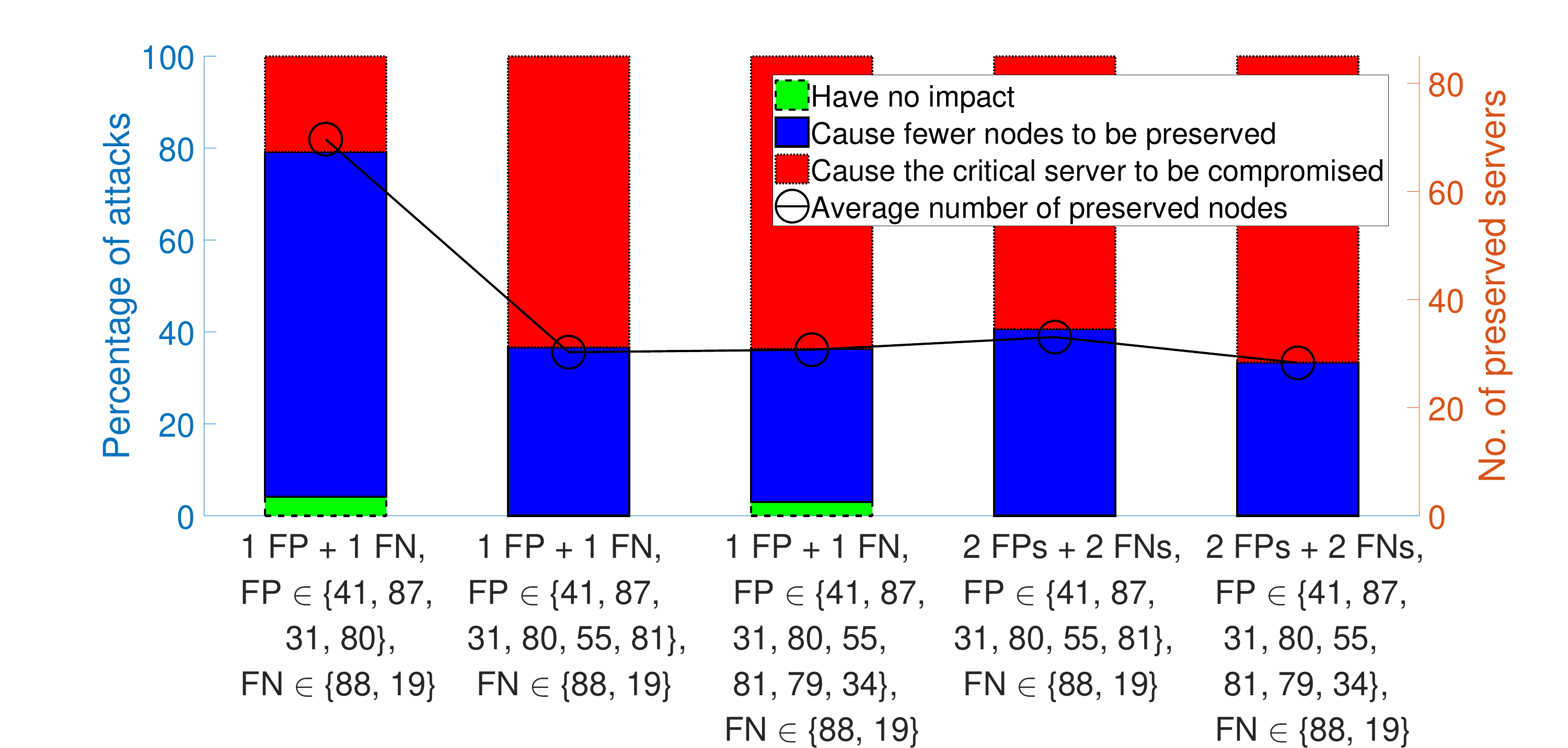}
  \caption{Setup 1: attacks against DDQN}
  \label{figure_attack_ddqn_36_46}
\end{subfigure}
\begin{subfigure}{.75\columnwidth}
  \centering
  \includegraphics[width=\columnwidth]{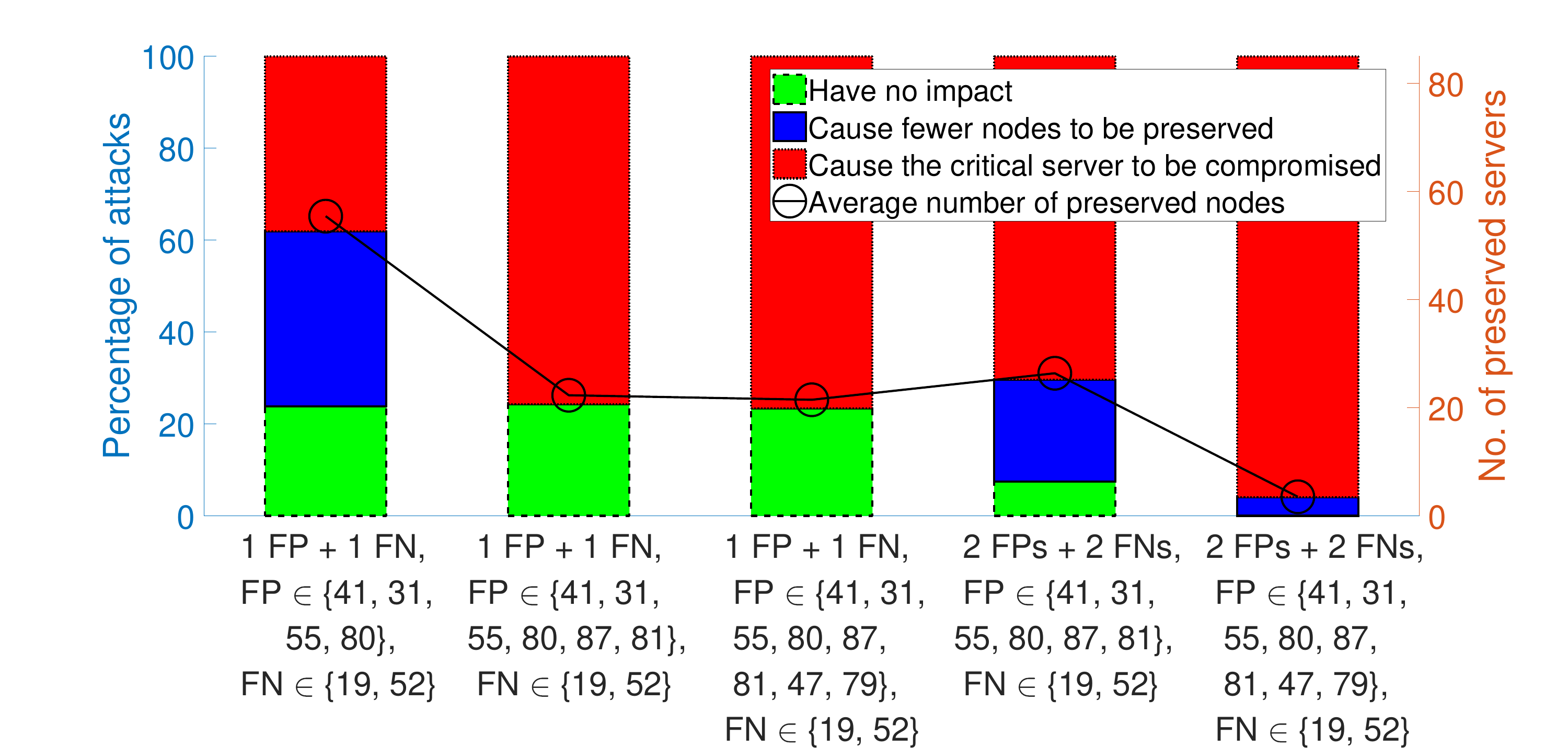}
  \caption{Setup 1: attacks against A3C}
  \label{figure_attack_a3c_34_46}
\end{subfigure}
\begin{subfigure}{.75\columnwidth}
  \centering
  \includegraphics[width=\columnwidth]{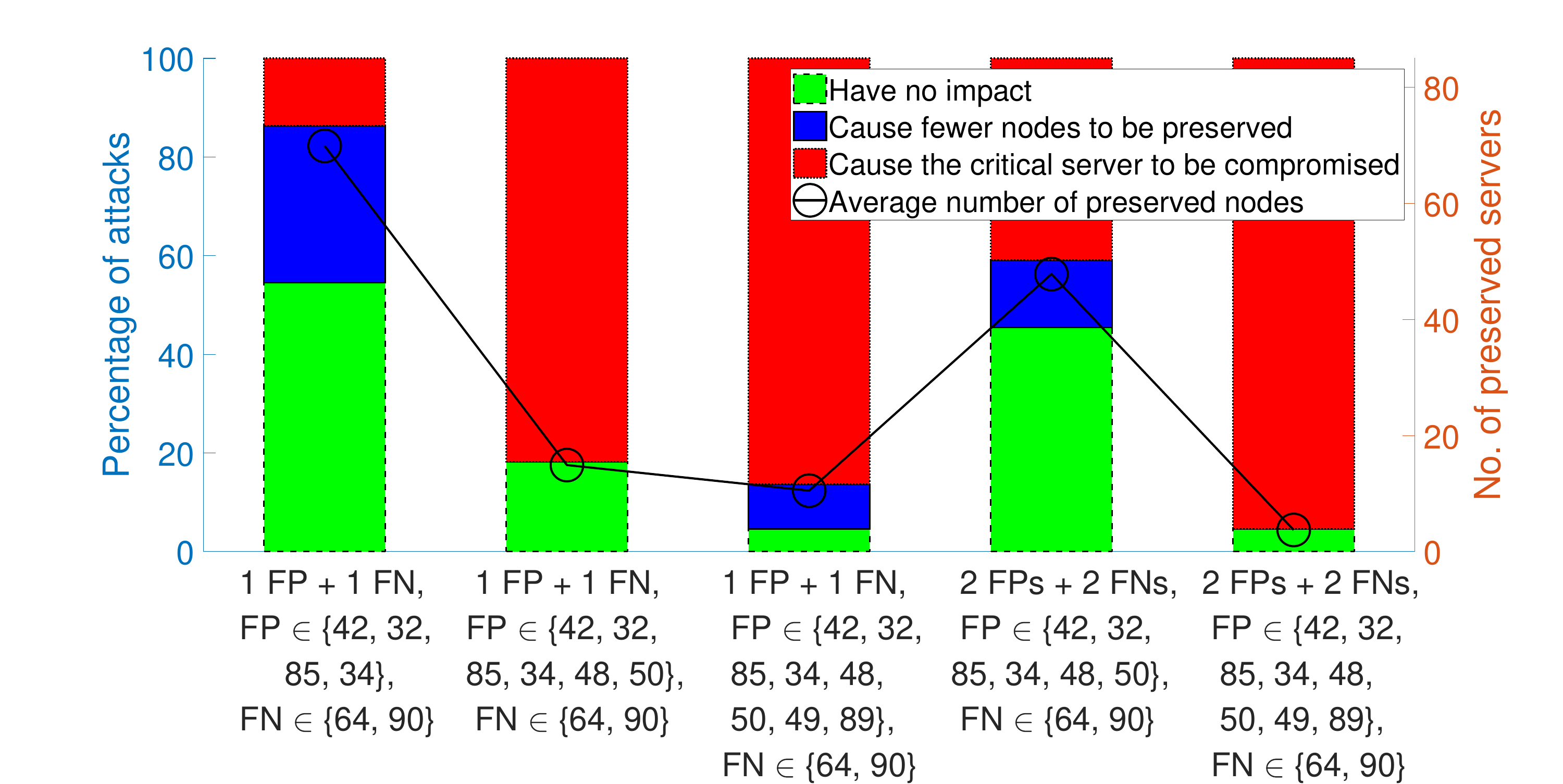}
  \caption{Setup 2: attacks against DDQN}
  \label{figure_attack_ddqn}
\end{subfigure}
\begin{subfigure}{.75\columnwidth}
  \centering
  \includegraphics[width=\columnwidth]{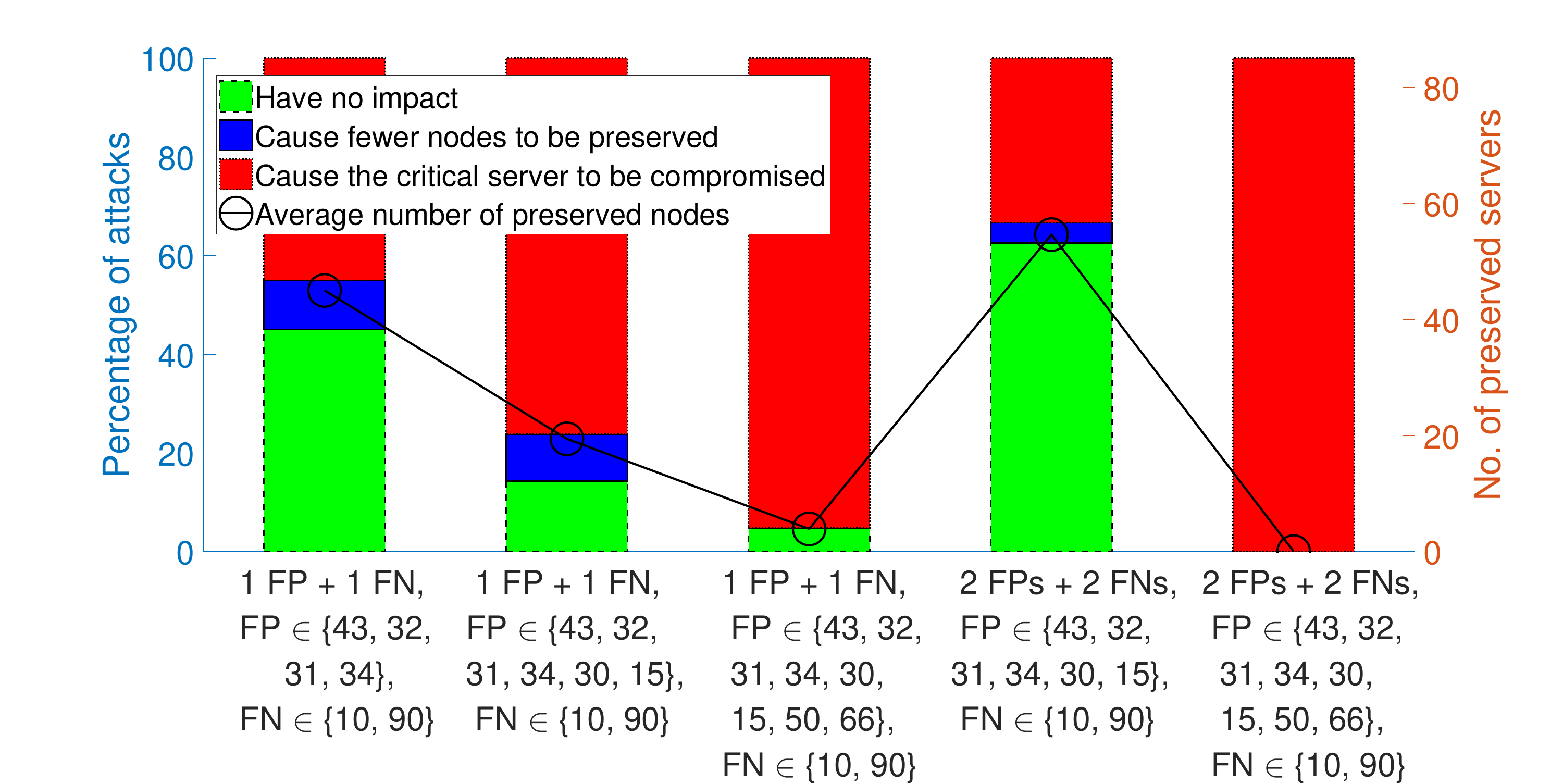}
  \caption{Setup 2: attacks against A3C}
  \label{figure_attack_a3c}
\end{subfigure}
\caption{Attacks against the DDQN \& A3C agents. The bars indicates the percentage of attacks (left \(y-\)axis) that 
(1) have no impact; (2) cause fewer nodes to be preserved; and (3) cause the critical server to be compromised. 
The lines 
indicate the average number of preserved servers (right \(y-\)axis).}
\label{figure_attack}
\end{figure}

\subsection{Countermeasure}\label{sec:experiment_defence}
Our inversion defence method only assumes that attackers perturb the states of a certain number of nodes in each training step, and aims to identify \& revert the manipulations. However, the defender has to loop through all the nodes rather than the nodes in \(L_\mathit{FP}\) \& \(L_\mathit{FN}\), and has to estimate the number of false readings added per step.

Specifically, four scenarios are investigated under the second setup: in the first three scenarios, the attacker adds two FPs \& two FNs per training step, and \(|L_\mathit{FP}|=8\), \(|L_\mathit{FN}|=2\) (\ie the most powerful form of attack studied in the experiments), while the defender assumes that there are (1) one FP \& one FN, (2) two FPs \& two FNs, (3) three FPs \& three FNs per training step. In the last case, the defender assumes that two FPs \& two FNs are added per time step, but in fact there is no attack. The first three scenarios investigate the situations where the defender either does or does not know the limit on the number of false readings added per time, while the last scenario is designed to study whether the normal learning process will be impacted when the defender falsely assumes the presence of an attack.

\begin{figure}[t!]
\centering
\begin{subfigure}{.7\columnwidth}
  \centering
  \includegraphics[width=\columnwidth]{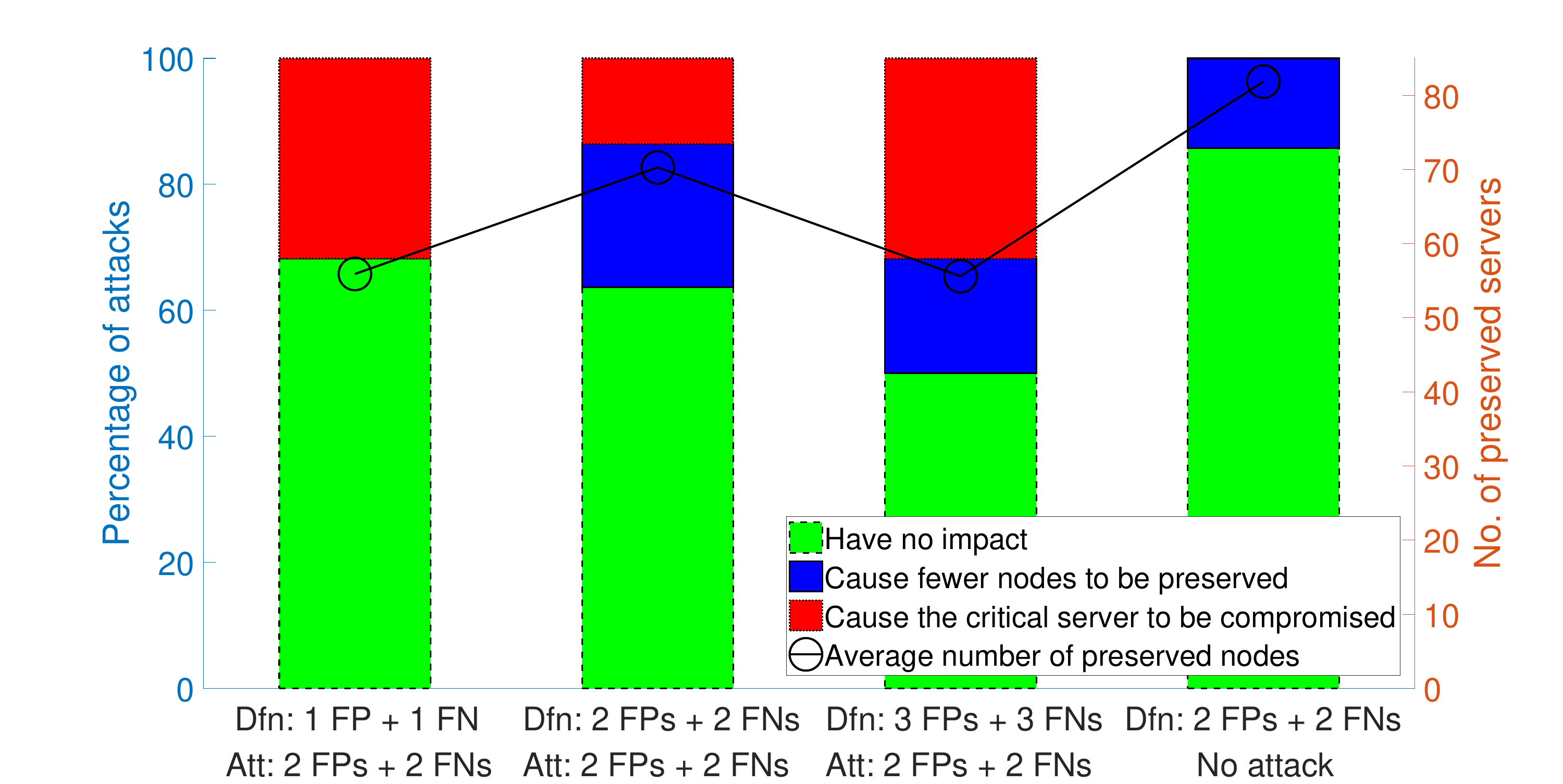}
  \caption{Defence against attacks on DDQN}
  \label{figure_defence_ddqn}
\end{subfigure}
\begin{subfigure}{.7\columnwidth}
  \centering
  \includegraphics[width=\columnwidth]{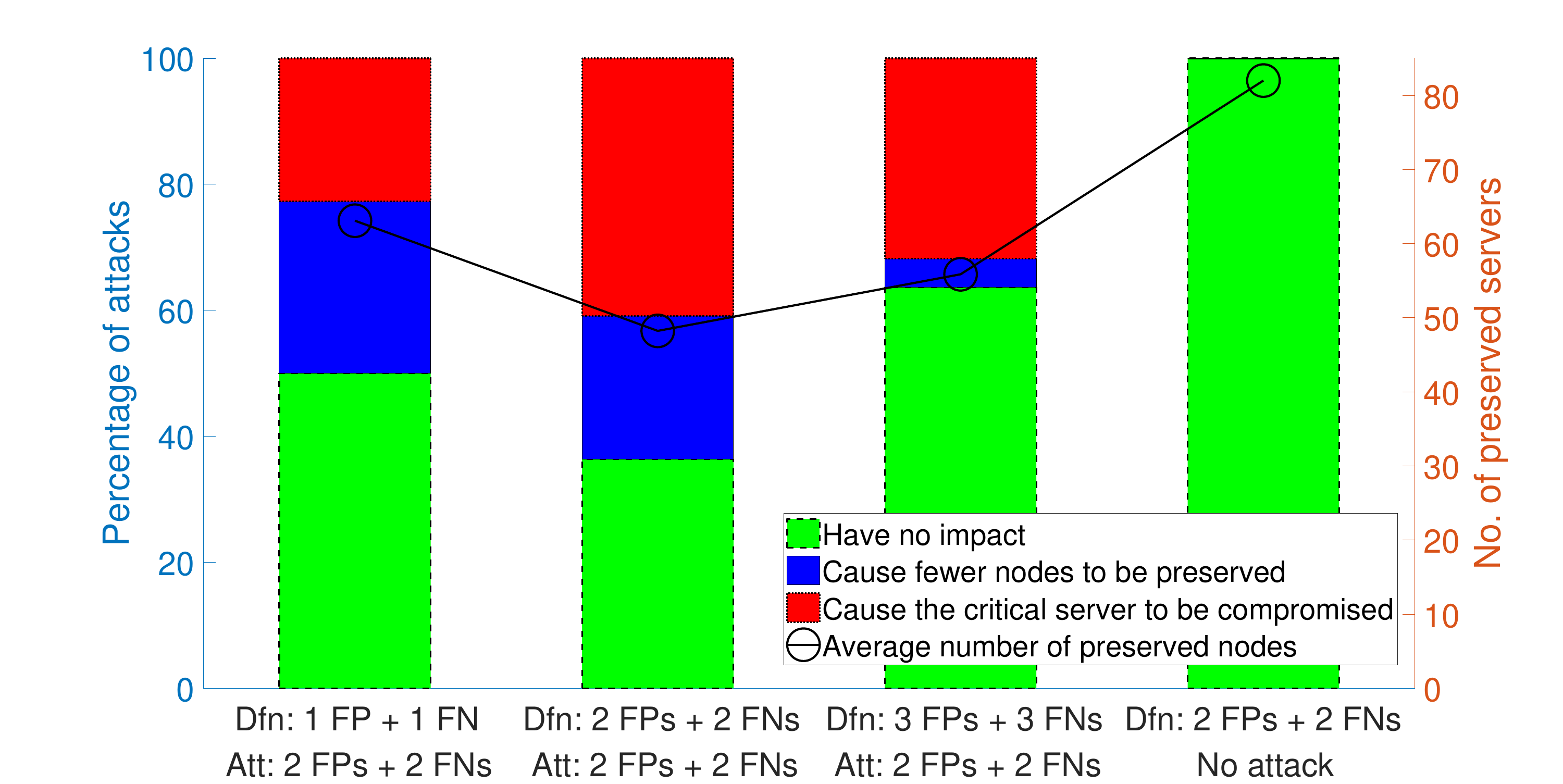}
  \caption{Defence against attacks on A3C}
  \label{figure_defence_a3c}
\end{subfigure}
\caption{Defence against attacks on the DDQN \& A3C agents.}
\label{figure_defence}
\end{figure}

Comparing the rightmost bars in Figs.~\ref{figure_attack_ddqn} \&~\ref{figure_attack_a3c} and the left three bars in Figs.~\ref{figure_defence_ddqn} \&~\ref{figure_defence_a3c}, we can see that the proposed defence method can effectively mitigate the impact of the causative attacks---the percentage of experiments where the critical server is compromised drops from almost 100\% to less than 30\% on average. In addition, the two rightmost bars in Fig.~\ref{figure_defence} also indicate that in most cases the defence method will not prevent the agent from learning the optimal actions when there is no attack---in all the cases represented by the blue bar in Fig.~\ref{figure_defence_ddqn}, only one less node is preserved.

\subsubsection{Discussion on the Overhead}\label{subsec:discussion}
A disadvantage of the inversion defence method is that it significantly slows down the training process, as it is time-consuming to loop through all the nodes to find the potential FPs and FNs. We aim to improve the performance in our future work. Specifically, we find that not all nodes are equally important in terms of preventing the critical server from being compromised---incorrect readings from certain nodes can cause more damage. Therefore, we will be investigating improving the efficiency of the defence method by only looping through those crucial nodes.

\section{Related Work}\label{sec:related}
This section first summarises adversarial machine learning against supervised classifiers, and then reviews recent work on similar attacks against reinforcement learning models. Finally, we discuss existing defence mechanisms.

\subsection{Adversarial Machine Learning}
Adversarial machine learning aims to minimise the modifications to the input, \ie either the test instance or the training sample, to cause a malfunction of the machine learning model.

Biggio \etal~\cite{biggio_poisoning_2012,biggio_security_2014} formulate the problem of evading a machine learning classifier as optimisation of the model's continuous scores, and use gradient descent to generate adversarial samples. Szegedy \etal~\cite{szegedy_intriguing_2013} highlight the observation that modifications imperceptible to humans can cause deep neural networks to misclassify, and they design the Fast Gradient Sign Method~\cite{goodfellow_explaining_2014} for the attack. Since then a number of different methods for creating adversarial samples have been proposed~\cite{nguyen_deep_2015,papernot_limitations_2016,papernot_transferability_2016,papernot_practical_2017,moosavi-dezfooli_universal_2016,moosavi-dezfooli_deepfool:_2016,carlini_towards_2016,madry_towards_2017}, among which the C\&W attack~\cite{carlini_towards_2016} is empirically the most efficient exploratory attack so far. In addition, more recent work has also studied adversarial attacks in other domains, such as graph-based models~\cite{zugner_adversarial_2018,dai_adversarial_2018}.

\subsection{Adversarial Reinforcement Learning}
It has been shown that reinforcement learning models are also vulnerable to the above attacks against classifiers. For example, Huang \etal \cite{huang_adversarial_2017} demonstrate that both white-box and black-box attacks using the Fast Gradient Sign Method~\cite{goodfellow_explaining_2014} are effective against deep RL.

Behzadan \& Munir \cite{behzadan_vulnerability_2017} were the first to investigate causative attacks against RL agents. They show how adversaries can perturb the observed state, in order to prevent the DQN agent from learning the correct policy.

Lin \etal \cite{lin_tactics_2017} propose two types of attacks against deep RL: (1) strategically-timed attack, which aims to decrease the number of time steps to launch the attack; (2) enchanting attack, which aims at misleading the agent to a specific state.

Pattanaik \etal~\cite{pattanaik_robust_2017} show that even the na\"ive attack, that is, adding random noise into the current state, is effective against deep RL---this is contrary to our experimental findings. However, our scenario is different to that described by the authors, including the dimensions of the state, the action space, and they design a gradient based attack that aims to maximise the probability of taking the worst possible action. 

\subsection{Existing Defence Mechanisms}
Generally speaking, existing defence methods against adversarial machine learning can be categorised into two classes: (1) data-driven defence, which either filters adversarial samples, injects adversarial samples into training---a.k.a., adversarial training, or projects inputs into a lower dimension; (2) learner robustification, which stabilises the training, applies moving target, or leverages ideas from robust statistics.

Countermeasures against attacks on RL models adopt similar approaches. Mandlekar \etal~\cite{mandlekar_adversarially_2017}, Pattanaik \etal~\cite{pattanaik_robust_2017} propose different adversarial training algorithms. 
Lin \etal~\cite{lin_detecting_2017} use previous images to predict future input and detect adversarial examples. Havens \etal~\cite{havens_online_2018} propose the Meta-Learned Advantage Hierarchy framework that measures the underlying changes in a task to detect the attack. Another line of work initiates the study of formal verification of deep RL~\cite{kazak_verifying_2019}.

\section{Conclusions and Future Work}\label{sec:conc}
In this paper, we show that in the context of autonomous defence in cyber networks, RL agents can be manipulated by attacks that target the training process, even if the attacker only has partial observability of the environment and defensive algorithms. In order to defend against the attack, we propose an inversion method that aims to revert the perturbations added by the attacker. Our experimental results demonstrate the effectiveness of the proposed approach, and show that it causes limited impact in non-attack scenarios. Our work focuses on learning in software-defined networking, which brings with it novel threat models of independent interest to adversarial learning research.

For future work, we plan to work on three directions---(1) partial observability: (i) impose partial observability also on the defender, as the defender may not obtain the correct states of all the nodes all the time; (ii) identify the minimum set of nodes the attacker needs to control for a certain level of effectiveness. (2) Consider a more powerful attacker that can (i) expand their partial observability as the attack proceeds; and (ii) spread more freely through the network, instead of having to compromise all the nodes on the paths to the critical server. (3) Replace the binary state with a continuous state.

\section{Acknowledgements}
This work was supported by the DST Group Next Generation Technologies Fund (Cyber) program via Data61 CRP `Adversarial Machine Learning for Cyber'.

\bibliographystyle{IEEEtran}
\bibliography{references}


\end{document}